\documentclass{article} 
\usepackage{iclr2025_conference,times}

\usepackage{amsmath,amsfonts,bm}

\def\eqref#1{equation~\ref{#1}}

\def\1{\bm{1}}

\DeclareMathAlphabet{\mathsfit}{\encodingdefault}{\sfdefault}{m}{sl}
\SetMathAlphabet{\mathsfit}{bold}{\encodingdefault}{\sfdefault}{bx}{n}

\usepackage{hyperref}
\usepackage{url}
\usepackage{booktabs}       
\usepackage{graphicx}

\title{Beyond Toy Benchmarks: \\ A Systematic Evaluation of OOD Detection Methods for Plant Pathology Classification}

\author{Devesh Shah \\
Independent Researcher \\
\textit{deveshs@umich.edu} \\
}

\iclrfinalcopy 
\begin{document}

\maketitle

\begin{abstract}
Out-of-distribution (OOD) detection is essential for reliable deployment of deep learning systems, yet the majority of existing methods are evaluated on small, visually homogeneous benchmarks. In this work, we study six OOD detection methods spanning post-hoc scoring, auxiliary objectives, energy-based models, and constrained optimization on the Plant Pathology 2021 dataset, a fine-grained task with natural distribution shifts. Energy-based fine-tuning performs best across OOD settings, improving detection over the softmax baseline while preserving in-distribution accuracy. Analysis shows these gains stem from both a restructuring of the embedding space alongside calibration of the scoring function. We further document practical training instabilities that arise when scaling constrained optimization methods to moderate-sized datasets, findings that are largely absent from existing literature. Our results demonstrate that principled OOD detection is achievable on real-world domain-specific data and that benchmark evaluations alone may not capture the challenges that emerge in practice.
\end{abstract}

\section{Introduction}
\label{sec:intro}

The past decade has seen remarkable progress in deep learning for computer vision, with models exceeding human-level performance on a range of benchmark tasks \citep{he2016deep, khan2022transformers}. However, this success relies on a key assumption: that training and deployment data are drawn from the same distribution. In practice, this assumption is frequently violated.

A substantial body of work has emerged to address out-of-distribution (OOD) detection; however, in practical deployment settings, deep learning systems remain largely rooted in a closed-world assumption, where models are not designed to express uncertainty or say "I don't know" \citep{amodei2016concrete}. This is in part because standard classification pipelines are trained using softmax-based objectives, which optimize for correct in-distribution discrimination rather than OOD awareness. This limitation is structural: softmax normalization forces probability mass to be distributed across known classes, causing models to assign high confidence to unrelated inputs \citep{nguyen2015deep}.

Most existing OOD detection methods today are evaluated on small, visually homogeneous benchmarks that do not reflect real-world deployment complexity \citep{yang2022openood}. This raises the question of how these methods behave under more realistic domain conditions, where semantic variation, covariate shift, and near-OOD cases are common. To address this gap, we study OOD detection in automated foliar disease classification in apple orchards. This setting provides a useful testbed because in-distribution performance is already strong \citep{plant-pathology-2021-fgvc8}, enabling OOD behavior to be analyzed in isolation. At the same time, deployment in agricultural environments introduces a diverse range of distribution shifts, including non-plant objects, inputs from different crops, visually similar but semantically distinct categories such as flowers, and environmental occlusions.

We conduct a systematic evaluation of OOD detection methods on the Plant Pathology 2021 dataset \citep{plant-pathology-2021-fgvc8} under varying assumptions about access to OOD data during training, emphasizing near-OOD performance as the most practically challenging regime. Code for all experiments is publicly available at \url{https://github.com/deveshshah1/Out_Of_Distribution_Exploration.git}.
\section{Background \& Related Work}

The softmax confidence score has become a common baseline for OOD detection \citep{hendrycks2016baseline}. A theoretical investigation \citep{hein2019relu} shows that neural networks with ReLU activation can produce arbitrarily high softmax confidence for OOD inputs. Subsequent post-hoc methods sought to improve this signal without retraining. \citet{liang2017enhancing} (ODIN) demonstrated that input preprocessing with adversarial perturbations and temperature scaling could make the softmax distribution more discriminative between in- and out-of-distribution examples. \citet{lee2018simple} proposed using class-conditional Gaussian distributions fitted to intermediate feature representations as an alternative scoring mechanism. While effective, these methods share a common limitation: the OOD score is derived entirely from a model trained without any OOD awareness, placing a ceiling on detection performance.

A parallel line of work introduced auxiliary OOD data during training to explicitly regularize model behavior on anomalous inputs. \citet{lee2017training} trained a classifier concurrently with a GAN, using generated samples as surrogate OOD data to suppress classifier confidence. Generative modeling offers a related alternative: rather than repurposing a discriminative classifier, one can train a model to explicitly estimate data density and flag low-density inputs as OOD \citep{grathwohl2019your}. However, \citet{nalisnick2018deep} showed that deep generative models can assign anomalously high likelihood to OOD data, undermining this intuition. Furthermore, generative approaches require training an additional network, making them substantially more expensive than discriminative methods on complex real-world data.

In this work, we focus on a complementary set of approaches that operate by fine-tuning an existing classifier rather than training auxiliary networks. \citet{hendrycks2018deep} proposed Outlier Exposure (OE), which uses a large, diverse dataset of real images disjoint from the test OOD distribution, training the model to produce uniform distributions on auxiliary OOD samples. \citet{liu2020energy} reframed the scoring problem in terms of the free energy of the classifier's logits rather than the softmax probability, preserving absolute logit magnitude information that normalization discards, and showed this can be further improved via fine-tuning with a margin-based energy shaping loss. More recent work has moved toward settings where clean auxiliary OOD data is unavailable entirely. \citet{katz2022training} (WOODS) proposed framing OOD detection as a constrained optimization problem over unlabeled \textit{wild} data, a mixture of in-distribution and OOD samples with no labels for either component, a more realistic assumption for systems that naturally accumulate unlabeled data of unknown provenance. \citet{bai2023feed} (SCONE) extended WOODS by incorporating covariate-shifted in-distribution data into the wild mixture, introducing an energy margin that prevents the model from rejecting semantically valid but visually degraded inputs.

The majority of OOD detection literature is benchmarked on small, visually homogeneous datasets such as CIFAR-10 and CIFAR-100, which may not reflect real-world deployment complexity \citep{yang2022openood}. Near-OOD detection, where inputs are visually similar to in-distribution data, remains substantially harder than far-OOD detection \citep{yang2024generalized} but is underrepresented in standard benchmarks. Furthermore, applied evaluations in domain-specific settings remain sparse; agricultural applications in particular face a wide range of semantic and covariate shift at deployment such as different crops, lighting conditions, occlusions, and visually similar but botanically distinct inputs such as healthy flowers. We address this gap by progressively evaluating several OOD detection methods on the Plant Pathology 2021 dataset \citep{plant-pathology-2021-fgvc8}, treating each method as a step in a deliberate assumption-relaxation sequence and evaluating across OOD sets of varying semantic distance from the training distribution.

\section{Methods}

\subsection{Dataset}

\subsubsection*{In-Distribution Data}
We evaluate on the Plant Pathology 2021 dataset \citep{plant-pathology-2021-fgvc8}, a real-world agricultural imaging dataset originally released as the FGVC8 Kaggle competition. The task involves classifying apple leaf images into foliar disease categories — a practically motivated problem, as foliar diseases pose a significant threat to orchard productivity and current diagnosis relies on labor-intensive manual scouting. The dataset presents natural challenges common to applied settings: class imbalance, limited total size, and the use of narrow domain-specific imagery. Rather than optimizing for competition performance, a problem largely considered solved, we repurpose this dataset as a testbed for studying OOD detection under realistic conditions.

\begin{figure}[tbp]
    \centering
    \includegraphics[width=\textwidth]{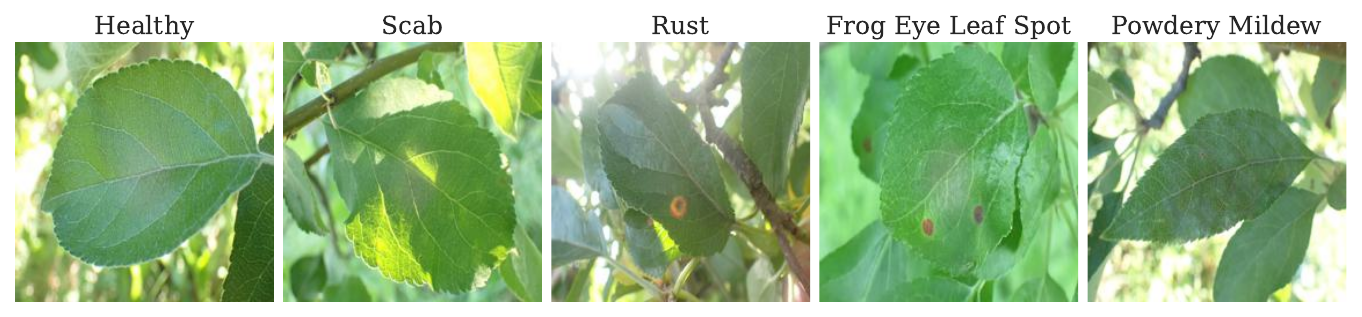}
    \caption{Representative examples of each disease class in the Plant Pathology 
    2021 dataset.}
    \label{fig:class_examples}
\end{figure}

The original dataset contains images with multiple labels and several low-frequency classes. To simplify the problem and focus on OOD detection rather than multi-label classification, we filter to the five most populous single-label classes, yielding a 15,675-image, 5-class classification problem. Multi-label OOD detection is left as future work, as overlapping class distributions substantially complicate the definition of in- versus out-of-distribution, a topic that is out of scope for this work. In the absence of provided splits, we assume a random stratified split is appropriate. The resulting split allocates 64\% to training, 16\% to validation, and 20\% to test, with an additional 15\% of the training pool held out as unlabeled wild data for use in later experiments. The full data distribution across splits and classes is shown in Table \ref{tab:id_dataset_distribution}. Representative examples of each class are visualized in Figure~\ref{fig:class_examples}, with additional per-class samples provided in Appendix~\ref{app:dataset}.

\begin{table}[htbp]
    \centering
    \caption{Number of images per disease class across dataset splits. Row labels indicate disease categories; values indicate image counts.}
    \label{tab:id_dataset_distribution}
    \begin{tabular}{lccccc}
        \toprule
         & Train & Train as Wild & Val & Test & Entire Dataset \\
        \midrule
        scab & 2,626 & 463 & 772 & 965 & 4,826 (31\%) \\
        healthy & 2,515 & 444 & 740 & 925 & 4,624 (29\%) \\
        frog\_eye\_leaf\_spot & 1,731 & 305 & 509 & 636 & 3,181 (20\%) \\
        rust & 1,011 & 179 & 298 & 372 & 1,860 (12\%) \\
        powdery\_mildew & 644 & 114 & 189 & 237 & 1,184 (8\%) \\
        \midrule
        Total & 8,527 (54\%) & 1,505 (10\%) & 2,508 (16\%) & 3,135 (20\%) & 15,675 \\
        \bottomrule
    \end{tabular}
\end{table}

\subsubsection*{Out-Of-Distribution Data}

We evaluate OOD detection performance across the following three external datasets of varying semantic distance from the training domain, none of which are seen during training. Representative examples of each are seen in Appendix~\ref{app:ood_dataset}.

\renewenvironment{quote}
  {\list{}{\leftmargin=1em\rightmargin=0pt\parsep=6pt}\item[]\relax}
  {\endlist}

\begin{quote}
    \textbf{Stanford Cars} \citep{Krause_2013_ICCV_Workshops} contains 8,041 test-set images of vehicle classes. As a far-OOD set with no visual overlap with plant imagery, this represents the easiest detection scenario.
    
    \textbf{Describable Textures Dataset (DTD)} \citep{cimpoi14describing} contains 5,640 images of visual textures across all splits. Textures are commonly used in OOD benchmarking as they differ structurally from object-centric training distributions, representing another far-OOD scenario. 
    
    \textbf{Flowers102} \citep{nilsback2008automated} contains 8,189 images of flower species across all splits. As the most visually similar dataset to plant leaf imagery, this represents our near-OOD evaluation set and the most practically relevant failure mode — a model that cannot distinguish flowers from diseased leaves would be unreliable in real agricultural deployment.
\end{quote}

For experiments requiring auxiliary OOD data during training, we use ImageNet-O \citep{hendrycks2021nae}, a curated set of 2,000 images from classes absent in ImageNet-1K, specifically selected because they elicit high-confidence incorrect predictions from ImageNet-trained classifiers. We split ImageNet-O into a training partition of 1,600 images (80\%) and a validation partition of 400 images (20\%).

\subsection{Experimental Setup}
\subsubsection*{Notation}
We define the following dataset notation used throughout the remainder of this paper:

\begin{itemize}
    \item $\mathcal{D}^{train}_{id}$: in-distribution training split from the plant classification dataset.
    \item $\mathcal{D}^{train}_{ood}$: training partition of ImageNet-O (1,600 images). No class labels are used; all samples are treated as OOD.
    \item $\mathcal{D}^{train}_{wild}$: union of the train wild split (Table~\ref{tab:id_dataset_distribution}) and $\mathcal{D}^{train}_{ood}$. All labels are discarded and every sample is assigned the label \textit{wild}. This dataset is approximately balanced between ID and OOD samples.
    \item $\mathcal{D}^{val}_{id}$: in-distribution validation split from the plant classification dataset.
    \item $\mathcal{D}^{val}_{ood}$: validation partition of ImageNet-O (400 images). All samples are treated as OOD.
    \item $\mathcal{D}^{test}_{id}$: in-distribution test split from the plant classification dataset, with ground-truth class labels.
\end{itemize}

OOD test sets are referred to by their names directly: Stanford Cars, DTD, and Flowers102.

\subsubsection*{Base Model}
All experiments build on a pretrained ResNet-18 backbone (ImageNet weights) with a 5-class classification head, fine-tuned on $\mathcal{D}^{train}_{id}$. This serves as our baseline. Unless otherwise noted in a method's experimental details, all methods share the same configuration; full details in Appendix~\ref{apdx:training_config}.

\subsubsection*{Evaluation Metrics}
We evaluate each method along two axes. For in-distribution performance, we report balanced accuracy across the five plant disease classes. For OOD detection, we report AUROC and FPR95 (false positive rate at 95\% true positive rate) on each OOD test set. Higher balanced accuracy and AUROC are better; lower FPR95 is better. In addition to these quantitative metrics, we provide qualitative analysis via histograms of ID/OOD energy scores and softmax confidence scores.

\subsection{Detection Algorithms}
The experiments in this paper explore a range of OOD detection algorithms under progressively relaxed assumptions about what information is available at training time. The goal is not to demonstrate monotonic performance improvement, as each method operates under a different set of assumptions that may favor different deployment scenarios, but rather to provide a comprehensive characterization of how these methods behave on a real-world problem.

\subsubsection*{E1: Softmax Baseline}
Our baseline model has no awareness of the OOD task. A ResNet-18 backbone 
pretrained on ImageNet is fine-tuned on $\mathcal{D}^{\text{train}}_{\text{id}}$ 
using standard cross-entropy loss:

\begin{equation}
    \mathcal{L}_{\text{CE}}(y, \hat{p}) = -\sum_{c=1}^{C} y_c \log (\hat{p}_c)
\end{equation}

where $y_c$ is the ground-truth indicator for class $c$ and 
$\hat{p}_c = \text{softmax}(z)_c$ is the predicted probability for class $c$ 
given logits $z$. The model is selected at best validation balanced accuracy. 
At inference, the OOD score is defined as the negative maximum softmax probability:

\begin{equation}
    S(x) = -\max_{c} \hat{p}_c
\end{equation}

such that higher scores indicate more OOD-like inputs. This scoring function 
serves as the reference point for all subsequent methods.

\subsubsection*{E2: Independent Binary Classification}

Building on the baseline, we replace the single softmax head with five independent 
sigmoid outputs, each trained with binary cross-entropy (BCE) loss:

\begin{equation}
    \mathcal{L}_{\text{BCE}} = -\frac{1}{N}\sum_{i=1}^{N} 
    \left[ y_i \log \hat{p}_i + (1 - y_i) \log(1 - \hat{p}_i) \right]
\end{equation}

where $\hat{p}_i = \sigma(z_i)$ is the sigmoid output for class $i$. Each head 
independently asks whether the input belongs to its class, removing the zero-sum 
competition imposed by softmax normalization. Under softmax, high confidence on 
one class necessarily suppresses all others, which can inflate confidence on OOD 
inputs. Independent sigmoid outputs remove this coupling — it is possible for all 
class scores to fall below the decision threshold simultaneously. The OOD score 
is defined as $S(x) = 1 - \max_c \hat{p}_c$. No OOD data is used; training and 
validation follow the same splits as E1.

\subsubsection*{E3: Explicit OOD Class}

E3 is the first OOD-aware method. We augment the classification head with a sixth 
output representing an explicit OOD class, trained on 
$\mathcal{D}^{\text{train}}_{\text{id}} \cup \mathcal{D}^{\text{train}}_{\text{ood}}$ 
using BCE loss per class as in E2. The OOD score is the predicted probability of 
the sixth class. This approach represents an upper bound of sorts on what is 
achievable when the OOD distribution is known and labeled at training time — a 
strong but operationally demanding assumption. In practice, introducing an entirely 
new class into a fine-tuned model disrupts the learned feature space; we therefore 
re-initialize from pretrained ImageNet weights rather than fine-tuning from E1. 
Validation is performed on 
$\mathcal{D}^{\text{val}}_{\text{id}} \cup \mathcal{D}^{\text{val}}_{\text{ood}}$.

\subsubsection*{E4: Outlier Exposure}

Outlier Exposure (OE) \citep{hendrycks2018deep} fine-tunes an existing classifier 
with an auxiliary loss that encourages the model to produce uniform predictive 
distributions on OOD samples, without adding any new classes or altering the 
output space. We initialize from the E1 checkpoint and fine-tune with a 
substantially reduced learning rate. The general OE objective is:

\begin{equation}
    \min_{\theta} \; \mathbb{E}_{(x,y) \sim \mathcal{D}^{\text{train}}_{\text{id}}} 
    \left[ \mathcal{L}_{\text{CE}}(f(x), y) \right] + 
    \lambda \cdot \mathbb{E}_{x' \sim \mathcal{D}^{\text{train}}_{\text{ood}}} 
    \left[ \mathcal{L}_{\text{OE}}(f(x')) \right]
\end{equation}

where $\lambda$ balances the two objectives. The goal is for the model to learn 
heuristics that distinguish in- from out-of-distribution inputs in a way that 
generalizes to unseen OOD distributions at test time. For our setting, where the 
baseline OOD detector is maximum softmax probability, $\mathcal{L}_{\text{OE}}$ 
is defined as the cross-entropy between the model's predictive distribution and 
the uniform distribution over $C$ classes:

\begin{equation}
    \mathcal{L}_{\text{OE}}(f(x')) = -\frac{1}{C} \sum_{c=1}^{C} \log \hat{p}_c(x')
\end{equation}

This penalizes the model for assigning high confidence to any single class on OOD 
inputs, encouraging flat, uncertain predictions. Training uses $\mathcal{D}^{\text{train}}_{\text{id}}$ and 
$\mathcal{D}^{\text{train}}_{\text{ood}}$ simultaneously via two dataloaders, 
with the OOD dataloader cycled to match the length of the ID dataloader. The OOD 
score at inference is identical to E1: $S(x) = -\max_c \hat{p}_c$.

\subsubsection*{E5: Energy-Based OOD Detection}

\citet{liu2020energy} propose using the free energy of a classifier's logits as 
an OOD scoring function, grounded in the framework of energy-based models 
\citep{lecun2006tutorial}. The core idea is to define a scalar energy function 
$E(x) : \mathbb{R}^D \rightarrow \mathbb{R}$ that maps each input to a single 
non-probabilistic scalar. For a neural network $f(x)$ with logits $f_c(x)$ for 
class $c$, the free energy is defined as:

\begin{equation}
    E(x; f) = -T \cdot \log \sum_{c=1}^{C} e^{f_c(x) / T}
\end{equation}

where $T$ is a temperature parameter. Unlike the softmax probability, which 
normalizes away absolute logit magnitude, the free energy aggregates across all 
class logits and preserves information about the overall confidence of the model. 
Lower energy indicates more in-distribution-like inputs; higher energy indicates 
more OOD-like inputs. The OOD score is thus $S(x) = E(x; f)$.

\paragraph{E5a: Post-Hoc Energy Score.} The energy score can be applied directly 
to the E1 checkpoint without any retraining, simply by replacing the MSP scoring 
function with the free energy. This serves as a parameter-free improvement over 
the softmax baseline, requiring no additional data or optimization.

\paragraph{E5b: Energy Fine-Tuning.} While the post-hoc energy score improves 
over MSP, the energy gap between in- and out-of-distribution data may not be 
optimal for a model trained without OOD awareness. We therefore fine-tune the 
model with a learning objective that explicitly shapes the energy 
surface. The full objective is:

\begin{equation}
    \mathcal{L} = \mathcal{L}_{\text{CE}}(f(x), y) + \lambda \cdot 
    \mathcal{L}_{\text{energy}}
\end{equation}

where $\mathcal{L}_{\text{energy}}$ is defined as:

\begin{equation}
    \mathcal{L}_{\text{energy}} = 
    \mathbb{E}_{(x_{\text{in}}, y) \sim \mathcal{D}^{\text{train}}_{\text{id}}}
    \left[ \max(0, E(x_{\text{in}}) - m_{\text{in}}) \right]^2 +
    \mathbb{E}_{x_{\text{out}} \sim \mathcal{D}^{\text{train}}_{\text{ood}}}
    \left[ \max(0, m_{\text{out}} - E(x_{\text{out}})) \right]^2
\end{equation}

The two squared hinge loss terms penalize ID samples with energy above $m_{\text{in}}$ 
and OOD samples with energy below $m_{\text{out}}$, explicitly pushing ID inputs 
toward lower energy and OOD inputs toward higher energy. Rather than adopting the 
margin values from the original paper, we derive $m_{\text{in}}$ and $m_{\text{out}}$ empirically from the energy 
distribution of the E1 checkpoint on $\mathcal{D}^{\text{val}}_{\text{id}}$ and 
$\mathcal{D}^{\text{val}}_{\text{ood}}$ respectively:

\begin{equation}
    m_{\text{in}} = Q_{0.50}\left(E(\mathcal{D}^{\text{val}}_{\text{id}})\right), 
    \qquad 
    m_{\text{out}} = Q_{0.50}\left(E(\mathcal{D}^{\text{val}}_{\text{ood}})\right)
\end{equation}

where $Q_\alpha$ denotes the $\alpha$-quantile. This ensures the margins are 
calibrated to the actual energy scale of our model and dataset rather than 
transferred from a mismatched setting. Training uses 
$\mathcal{D}^{\text{train}}_{\text{id}}$ and $\mathcal{D}^{\text{train}}_{\text{ood}}$ 
simultaneously, initialized from the E1 checkpoint.

\subsubsection*{E6: WOODS}

\citet{katz2022training} introduce a framework for training OOD detectors using 
unlabeled \textit{wild} data — a naturally occurring mixture of in-distribution 
and OOD samples that accumulates whenever a classifier is deployed in a real-world 
system. Rather than assuming access to clean, labeled OOD data as in E3--E5, 
WOODS operates entirely on unlabeled data of unknown provenance, making it the 
most realistic assumption of any method in this study.

Our wild dataset is constructed as 
$\mathcal{D}^{\text{train}}_{\text{wild}} = \mathcal{D}^{\text{train}}_{\text{wild-pool}} 
\cup \mathcal{D}^{\text{train}}_{\text{ood}}$, where 
$\mathcal{D}^{\text{train}}_{\text{wild-pool}}$ consists of unlabeled plant images 
held out from training and $\mathcal{D}^{\text{train}}_{\text{ood}}$ is the 
ImageNet-O training partition. All labels are discarded. The ratio of ID to OOD 
samples within the wild set is an important factor in method performance — 
\citet{katz2022training} provide a full sensitivity analysis on this ratio; we 
assume an approximately equal mixture of 0.5.

The framework jointly learns an OOD detector $g_\theta$ and multi-class classifier 
$f_\theta$ by framing detection as a constrained optimization problem — minimizing 
the rate at which OOD wild samples are classified as in-distribution, subject to 
constraints on the ID false alarm rate and classification accuracy:

\begin{equation}
    \min_{\theta} \; \mathbb{E}_{x \sim P_{\text{wild}}} 
    \left[ \ell_{\text{ood}}(g_\theta(x), \text{in}) \right]
\end{equation}
\begin{equation}
    \text{subject to} \quad
    \mathbb{E}_{x \sim P_{\text{in}}} 
    \left[ \ell_{\text{ood}}(g_\theta(x), \text{out}) \right] \leq \alpha, \quad
    \mathbb{E}_{(x,y) \sim P_{XY}} 
    \left[ \ell_{\text{cls}}(f_\theta(x), y) \right] \leq \tau
\end{equation}

where $\alpha$ is the maximum tolerated false alarm rate on ID data and $\tau$ 
bounds the allowable classification loss. The constrained problem is solved via 
the Augmented Lagrangian Method (ALM), which introduces additional dual variables, 
penalty weights, and constraint satisfaction tolerances; we refer the reader to 
\citet{katz2022training} for the full derivation. In practice, these additional 
variables make WOODS considerably more sensitive to hyperparameter choices than 
prior methods — during our implementation we found that careful tuning of the 
dual learning rate, penalty growth factor, and penalty weight cap was essential 
for stable training, a finding we discuss further in Section~\ref{sec:discussion}.
We initialize from the E1 checkpoint and reset all ALM state variables before 
training.
\section{Results}
Table~\ref{tab:main_results} summarizes OOD detection performance across all six methods and three evaluation sets. E3 achieves near-perfect AUROC and FPR95 across all evaluation sets by a substantial margin. While the training OOD distribution in E3 is still disjoint from our evaluation sets, E3 benefits from training with the explicit assumption that OOD is a known and labeled concern from the start — an operationally demanding assumption that is rarely achievable in practice. We therefore treat it as an approximate upper bound rather than a fair comparison point for the remaining methods.

\begin{table}[tbp]
    \centering
    \caption{OOD detection results across all methods. E3 serves as an approximate upper bound given its access to labeled OOD data 
at training time and is excluded from comparison. \textbf{Bold} indicates the 
best result per column among all other methods.}
    \label{tab:main_results}
    \resizebox{\textwidth}{!}{%
    \begin{tabular}{lcccccccc}
        \toprule
        & \multicolumn{1}{c}{} & \multicolumn{3}{c}{AUROC $\uparrow$} & \multicolumn{3}{c}{FPR95 $\downarrow$} \\
        \cmidrule(lr){3-5} \cmidrule(lr){6-8}
        Method & Bal. Acc $\uparrow$ & Stanford Cars & DTD & Flowers102 & Stanford Cars & DTD & Flowers102 \\
        \midrule
        E1: Baseline Softmax      & 0.934 & 0.482 & 0.775 & 0.415 & 1.000 & 0.942 & 1.000 \\
        E2: Independent BCE       & \textbf{0.942} & 0.751 & 0.844 & 0.794 & 0.904 & 0.660 & 0.745 \\
        E3: OOD Class             & 0.944 & 1.000 & 1.000 & 0.999 & 0.000 & 0.000 & 0.001 \\
        E4: Outlier Exposure      & 0.936 & 0.873 & 0.933 & 0.826 & 0.551 & 0.276 & 0.847 \\
        E5a: Energy Score (Post-Hoc) & 0.934 & 0.516 & 0.777 & 0.404 & 0.995 & 0.956 & 0.996 \\
        E5b: Energy Fine-Tuning   & 0.923 & \textbf{0.938} & \textbf{0.942} & \textbf{0.952} & \textbf{0.306} & \textbf{0.231} & \textbf{0.212} \\
        E6: WOODS                 & 0.916 & 0.777 & 0.874 & 0.762 & 0.858 & 0.597 & 0.847 \\
        \bottomrule
    \end{tabular}%
    }
\end{table}

Across all methods, balanced accuracy on $\mathcal{D}^{\text{test}}_{\text{id}}$ remains remarkably stable, ranging from 0.916 to 0.944. Even WOODS which introduces a competing wild-data objective from the first epoch incurs only a 1.8\% drop relative to the baseline. In fact, in several methods accuracy improves slightly over E1, suggesting that OOD-aware training can act as a mild regularizer on the classification task. Among methods that do not assume labeled OOD data at initial training time, E5b (energy fine-tuning) achieves the strongest OOD detection performance across all three evaluation sets. Comparing E5a to E1 confirms that the energy score alone already provides a meaningful improvement over the softmax baseline, and fine-tuning on top of this further separates the energy distributions of ID and OOD data. It is also worth noting that E2 provides a substantial improvement over E1 despite requiring no OOD data whatsoever — simply replacing the softmax head with independent sigmoid outputs improves AUROC by 0.269 on Stanford Cars and 0.379 on Flowers102, suggesting this deserves consideration as a stronger default baseline in future work.

Contrary to our initial expectation, Stanford Cars and Flowers102 prove equally challenging across most methods, while DTD is consistently the easiest evaluation set. This is a notable finding: what a human observer would classify as far-OOD (vehicles) can be as difficult for a model to reject as visually similar near-OOD inputs (flowers), underscoring the gap between perceptual similarity and learned feature representations. Finally, WOODS shows a meaningful improvement over the softmax baseline despite operating on an aggressive ID-to-OOD wild ratio of 0.5. In practice, deployed systems are likely to encounter a higher proportion of OOD samples in accumulated wild data, which would be expected to further improve WOODS performance and potentially close the gap with E4 and E5b.

\begin{figure}[tbp]
    \centering
    \includegraphics[width=\textwidth]{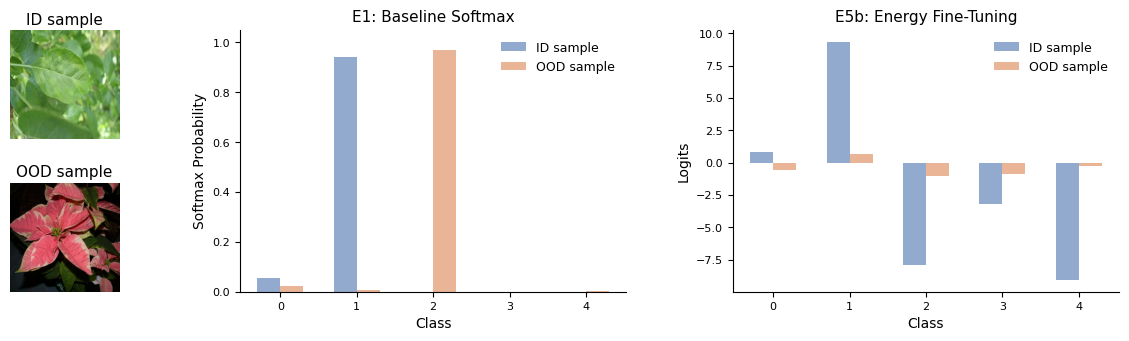}
    \caption{Softmax probabilities (E1) and logits (E5b) for a representative ID 
    sample (diseased plant leaf) and OOD sample (Flowers102). Under E1, the OOD 
    sample receives a near-certain prediction on class 2, illustrating the 
    overconfidence failure mode of softmax-based classifiers. Under E5b, the OOD 
    sample produces uniformly low logit magnitudes across all classes relative to 
    the ID sample, yielding energy scores of $-9.4$ (ID) and $-1.4$ (OOD) that 
    clearly distinguish the two inputs.}
    \label{fig:score_comparison}
\end{figure}

\begin{figure}[bp]
    \centering
    \includegraphics[width=\textwidth]{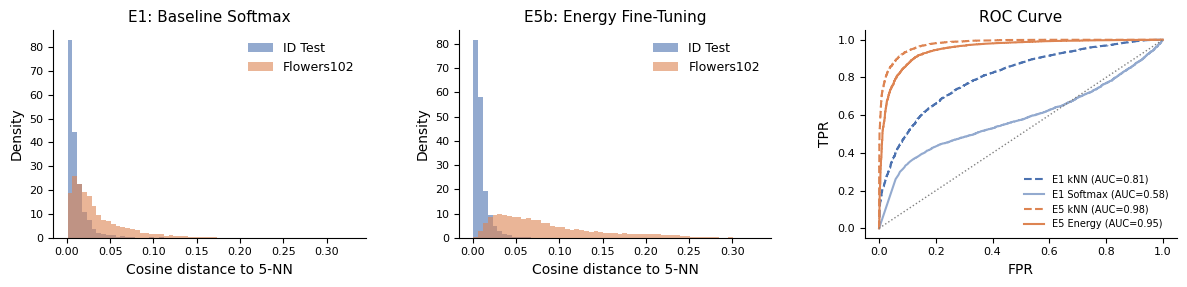}
    \caption{Left and center: distributions of cosine distance to the 5 nearest neighbors 
in the training ID embedding space, computed for 
$\mathcal{D}^{\text{test}}_{\text{id}}$ and Flowers102 under E1 and E5b 
respectively ($\mathcal{W} = 0.02$ and $\mathcal{W} = 0.07$). Right: ROC curves 
for softmax/energy-based and 5-NN cosine distance OOD scores on Flowers102 
for E1 and E5b.}
    \label{fig:knn_roc}
\end{figure}

We provide a qualitative assessment of results comparing E1 and E5b, the strongest 
performing method in our experiments. Figure~\ref{fig:score_comparison} visualizes 
the softmax probabilities and logits for a single ID and OOD sample pair. The OOD 
sample, a flower image from Flowers102, receives a near-certain prediction on 
class 2 under E1, illustrating the overconfidence failure mode described in 
Section~\ref{sec:intro}. E5b tells a markedly different story: the logit magnitudes 
for the OOD sample are uniformly low across all classes relative to the ID sample, 
making the two inputs far more distinguishable. This is directly captured by the 
energy score $-\log\sum_c e^{f_c(x)}$, which assigns $-9.4$ to the ID sample and 
$-1.4$ to the OOD sample, a clear separation that the softmax baseline entirely 
fails to produce.

To better understand what drives this improvement, we analyze the embedding space 
geometry rather than relying on 2D projections, which capture too little of the 
relevant variance to be informative for ID/OOD separation. For every test 
sample we compute the cosine distance to its 5 nearest neighbors in the training 
ID embedding space, and plot the resulting distributions for 
$\mathcal{D}^{\text{test}}_{\text{id}}$ and Flowers102 in 
Figure~\ref{fig:knn_roc}. Visually, E5b produces substantially greater separation 
between the two distributions compared to E1. Quantifying this with the Wasserstein 
distance, which measures the minimum cost of transforming one distribution into 
the other with larger values indicating greater distributional separation, yields 
$\mathcal{W} = 0.02$ for E1 and $\mathcal{W} = 0.07$ for E5b, confirming that 
energy fine-tuning meaningfully reshapes the embedding space.

Using the 5-NN cosine distance as a surrogate OOD score, we compute ROC curves 
for both methods on Flowers102, shown alongside the energy and softmax ROC curves 
in Figure~\ref{fig:knn_roc}. Two findings stand out. First, even under E1, the 
embedding distance achieves an AUROC of 0.81 compared to only 0.58 for the softmax 
score, suggesting that the baseline model's learned representations already contain 
meaningful ID/OOD signal that the softmax output fails to surface. Second, E5b 
improves both the energy-based AUROC (0.95) and the embedding distance AUROC 
(0.98). This suggests that energy fine-tuning improves OOD detection through at 
least two complementary mechanisms: a better-calibrated scoring function and a 
reorganization of the embedding space, with the latter appearing to contribute 
meaningfully given the gains observed in the kNN-based AUROC.

\subsection{Discussion}
\label{sec:discussion}

The central tradeoff across E1--E6 is between the realism of the auxiliary data 
assumption and detection performance. Methods that begin with more favorable 
assumptions about OOD data availability at training time achieve better results, 
yet such assumptions are rarely satisfied in practice. WOODS represents the most 
realistic scenario but pays for it with the highest training complexity of any 
method evaluated.

A consistent finding across E4, E5b, and E6 is that learning rate is the single 
most impactful hyperparameter for stable training. The OOD objective actively 
competes with the classification objective from the first epoch, and too high a 
learning rate allows the OOD loss to dominate before classification has stabilized. 
In E4, an aggressive $\lambda$ combined with a high LR caused the model to collapse 
toward uniform predictions on ID data; in E5b, it caused the energy margins to be 
violated immediately, preventing any meaningful energy gap from forming. 
Additionally, while \citet{hendrycks2018deep} recommend training from scratch with 
simultaneous OE loss, we found this did not converge stably at our dataset scale 
of 10K images. Fine-tuning from the E1 checkpoint with a reduced learning rate 
and a classification warmup period was necessary for stable convergence, suggesting 
that the from-scratch recommendation may not transfer to smaller datasets.

WOODS introduced additional instability that warrants particular attention for 
practitioners. The Augmented Lagrangian Method introduces dual variables 
$\lambda_1, \lambda_2$ and penalty weights $\beta_1, \beta_2$ that must be jointly 
tuned alongside the model's learning rate. In our initial experiments using 
$\eta_\lambda = 0.1$ and $\alpha = 0.05$, $\beta_1$ grew unboundedly to values 
exceeding 20, causing the wild-data loss term to dominate the entire objective, 
resulting in a 20\% collapse in validation accuracy and ID energy values shooting 
from $-8$ toward $0$. The root cause was twofold: the dual learning rate was too 
high, causing $\lambda$ to overshoot on every constraint violation, which in turn 
triggered $\beta$ growth at every epoch. The fix required reducing $\eta_\lambda$ 
from 0.1 to 0.001, loosening $\alpha$ from 0.05 to 0.1, and introducing a hard 
cap of $\beta_{\max} = 5.0$, a modification not present in the original WOODS 
formulation. We find that explicit $\beta$ clamping is a necessary stabilization 
measure when applying WOODS at moderate dataset scales, where $\beta$ growth is 
not naturally self-limiting as assumed by the original paper.

This study does not evaluate SCONE \citep{bai2023feed}, 
which extends WOODS by incorporating covariate-shifted in-distribution data into 
the wild mixture to improve robustness to visually degraded but semantically valid 
inputs. In our setting, the training augmentation pipeline already applies 
aggressive spatial and color transformations, and producing covariate-shifted 
variants beyond this regime distorted plant images beyond recognition, making a 
meaningful evaluation of SCONE infeasible.
\section{Conclusion}

We presented a systematic evaluation of six OOD detection methods on the Plant 
Pathology 2021 dataset, spanning a progression from simple post-hoc scoring to 
constrained wild-data optimization. Across all methods, in-distribution 
classification accuracy remained stable, confirming that OOD-aware training need 
not come at the cost of task performance. Energy-based fine-tuning (E5b) emerged 
as the strongest practical method, achieving the best OOD detection performance 
without requiring labeled OOD data or knowledge of the test distribution. Further 
qualitative analysis suggested its gains stem from both a reorganization of the 
embedding space and a recalibration of the scoring function.

Scaling these methods to a moderate-sized real-world dataset surfaced practical 
challenges that benchmark evaluations do not expose, particularly around training 
stability in constrained optimization methods, where careful tuning of dual 
learning rates and penalty weight caps proved essential. We hope these findings 
serve as a useful reference for practitioners applying OOD detection beyond 
standard benchmarks and motivate future work toward more robust and scalable 
wild-data methods for domain-specific deployment settings.

\bibliography{iclr2025_conference}
\bibliographystyle{iclr2025_conference}

\newpage
\appendix
\section{Training Configuration}
\label{apdx:training_config}

All experiments share a common base training configuration unless explicitly 
overridden in the method description. We document the full configuration here 
for reproducibility.

\renewenvironment{quote}
  {\list{}{\leftmargin=1em\rightmargin=0pt\parsep=6pt}\item[]\relax}
  {\endlist}

\begin{quote}

\textbf{Training augmentations:} Input images are resized to $224 \times 224$ and augmented with random horizontal and vertical flips, random rotation up to $20°$, Gaussian blur with kernel size $(5, 9)$ and sigma range $(0.1, 5)$, and color jitter applied to brightness, contrast, saturation, and hue.

\textbf{Class imbalance:} We apply inverse-frequency weighted random sampling during training, so underrepresented classes are sampled proportionally more often.

\textbf{Optimization:} We use AdamW with a learning rate of $1\times10^{-3}$ and weight decay of $1\times10^{-4}$, paired with a Noam scheduler with 4,000 warmup steps. Models are trained for 25 epochs with a batch size of 32.

\textbf{Checkpointing:} We save checkpoints at the best training loss, best validation loss, and best validation balanced accuracy, and select the best checkpoint across these three criteria for evaluation.

\textbf{Loss and metrics:} The baseline uses cross-entropy loss. We track accuracy and per-class balanced accuracy on both train and validation sets throughout training. Experiment tracking is done via Weights \& Biases (\href{https://wandb.ai/deveshshah-university-of-michigan/OOD_Exploration}{project workspace here}).

\textbf{Compute:} All experiments are trained on a single Apple M2 GPU. Due to compute constraints, we do not perform exhaustive hyperparameter searches; instead, we conduct small targeted sweeps per method and report results for the best-performing configuration found.

\end{quote}

\newpage
\section{ID Dataset Visualization}
\label{app:dataset}

The Plant Pathology 2021 dataset contains high-resolution RGB images of apple 
leaves captured under natural field conditions, exhibiting substantial visual 
variation within each class due to differences in lighting, background, leaf 
maturity, and imaging angle. After filtering to the five most populous single-label 
classes, the dataset covers a range of disease presentations that can be visually 
subtle and difficult to distinguish even for human observers. Figure~\ref{fig:class_examples} 
shows a single representative image per class, while Figure~\ref{fig:class_examples_appendix} 
below shows five randomly sampled images per class to illustrate the degree of 
intra-class variation present in the training data.

\begin{figure}[htbp]
    \centering
    \includegraphics[width=\textwidth]{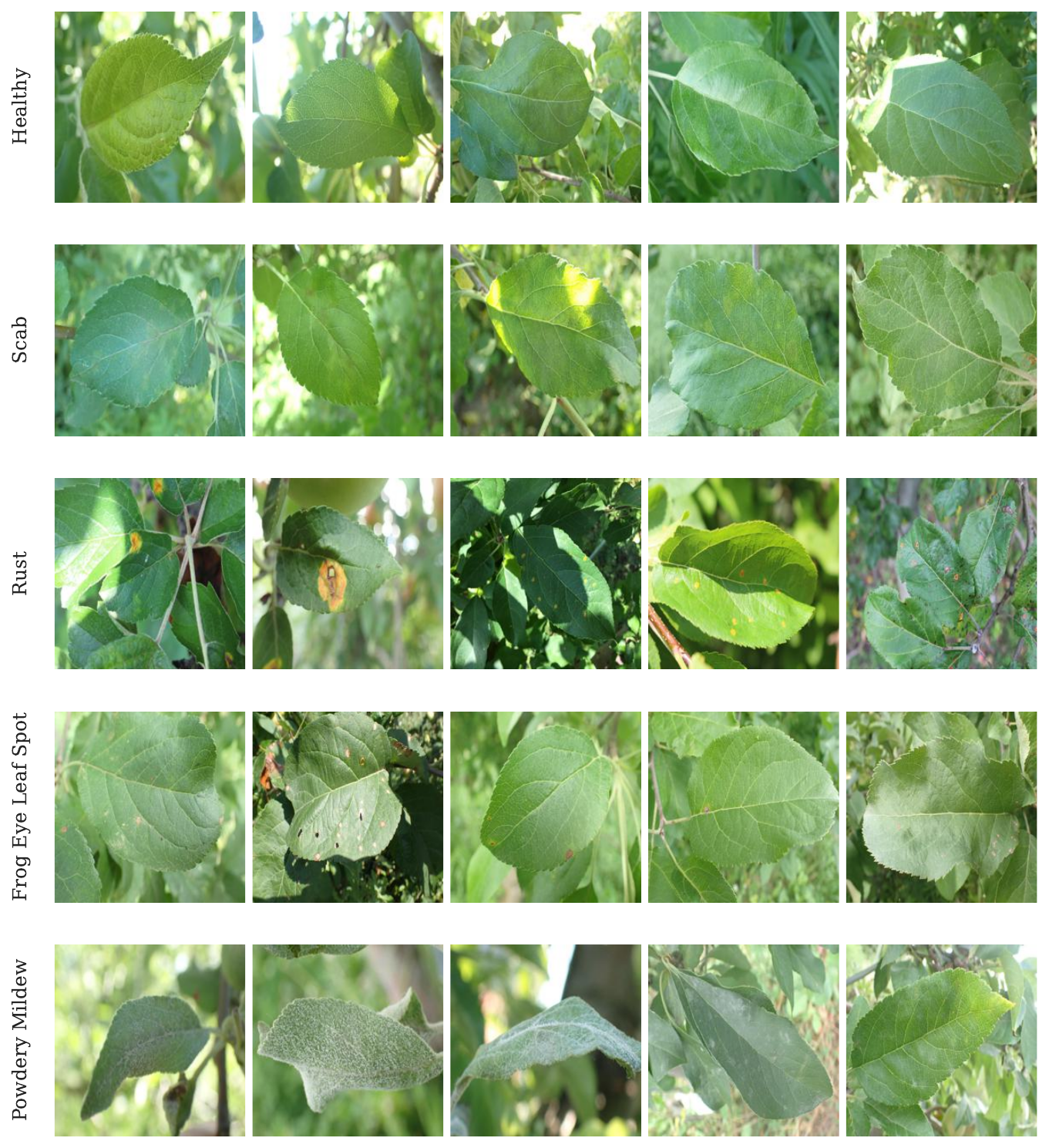}
    \caption{Five randomly sampled images per disease class from the training 
    split of the Plant Pathology 2021 dataset, illustrating the visual variation 
    within each class.}
    \label{fig:class_examples_appendix}
\end{figure}
\newpage
\section{OOD Dataset Visualization}
\label{app:ood_dataset}

\begin{figure}[htbp]
    \centering
    \includegraphics[width=0.8\textwidth]{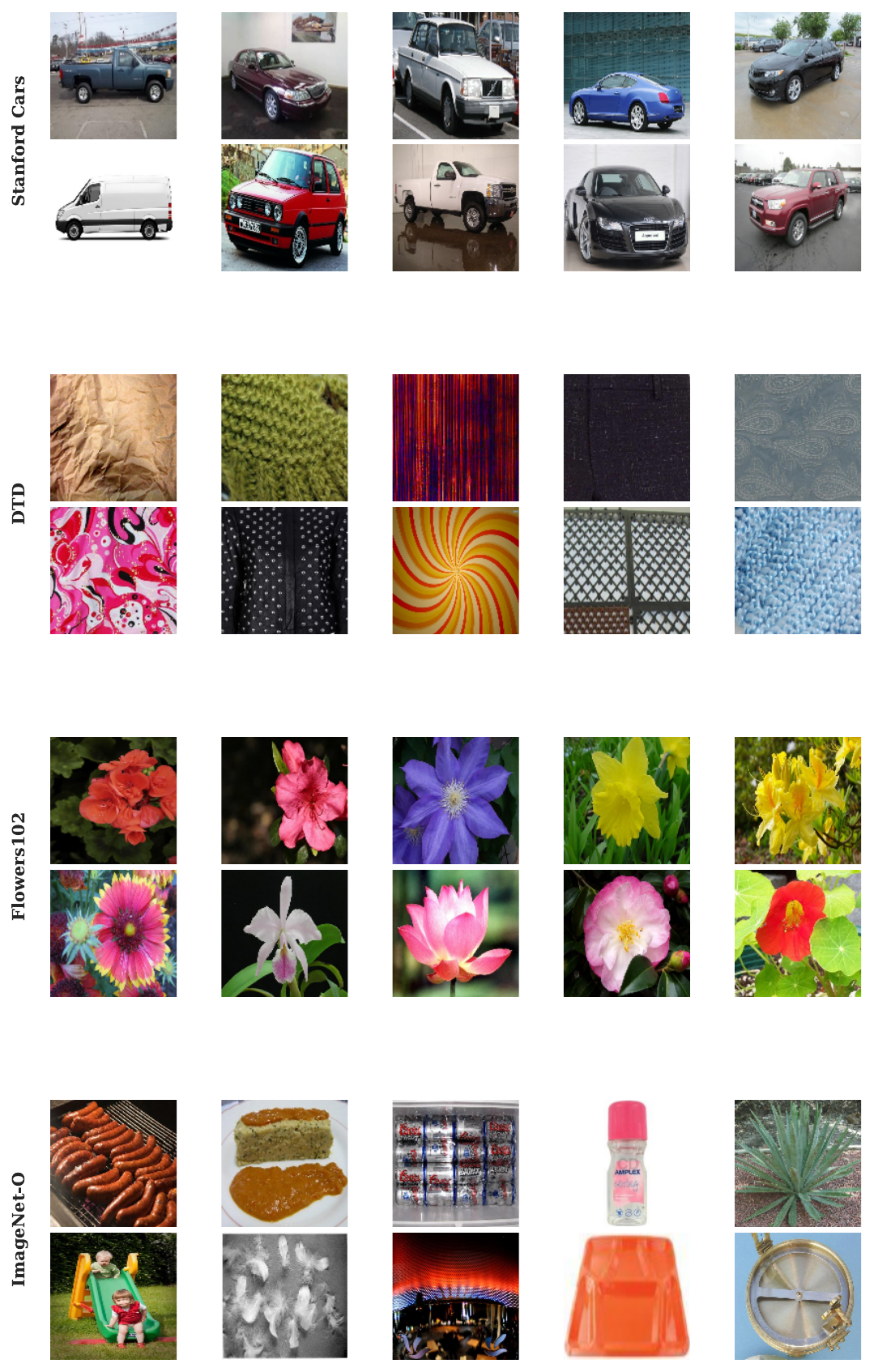}
    \caption{Randomly sampled images from each of the four OOD datasets used 
    in this study.}
    \label{fig:ood_examples}
\end{figure}

To provide intuition for the varying semantic distance between each OOD evaluation 
set and the in-distribution plant pathology data, we visualize representative 
examples from each of the four OOD datasets used in this study. Figure~\ref{fig:ood_examples} 
shows ten randomly sampled images from each dataset. Stanford Cars and DTD 
represent far-OOD scenarios with little visual overlap with plant imagery, while 
Flowers102 represents the most challenging near-OOD case given its visual 
similarity to leaf images. ImageNet-O, used only during training as auxiliary OOD 
data, contains diverse and visually complex images specifically selected to elicit 
high-confidence incorrect predictions from ImageNet-trained classifiers.

\end{document}